# Evaluating the Impact of Point Cloud Colorization on Semantic Segmentation Accuracy

Qinfeng Zhu
Department of Civil Engineering
Xi'an Jiaotong-Liverpool University
Suzhou, China
Qinfeng.Zhu21@student.xjtlu.edu.cn

Jiaze Cao
Department of Civil Engineering
Xi'an Jiaotong-Liverpool University
Suzhou, China
Jiaze.Cao19@alumni.xjtlu.edu.cn

Yuanzhi Cai
CSIRO Mineral Resources
CSIRO
Kensington, Australia
Yuanzhi.Cai@CSIRO.AU

Lei Fan*
Department of Civil Engineering
Xi'an Jiaotong-Liverpool University
Suzhou, China
Lei.Fan@xjtlu.edu.cn
*Corresponding Author

*Abstract*—Point cloud semantic segmentation, the process of classifying each point into predefined categories, is essential for 3D scene understanding. While image-based segmentation is widely adopted due to its maturity, methods relying solely on RGB information often suffer from degraded performance due to color inaccuracies. Recent advancements have incorporated additional features such as intensity and geometric information, yet RGB channels continue to negatively impact segmentation accuracy when errors in colorization occur. Despite this, previous studies have not rigorously quantified the effects of erroneous colorization on segmentation performance. In this paper, we propose a novel statistical approach to evaluate the impact of inaccurate RGB information on image-based point cloud segmentation. We categorize RGB inaccuracies into two types: incorrect color information and similar color information. Our results demonstrate that both types of color inaccuracies significantly degrade segmentation accuracy, with similar color errors particularly affecting the extraction of geometric features. These findings highlight the critical need to reassess the role of RGB information in point cloud segmentation and its implications for future algorithm design.

## I. INTRODUCTION

Semantic segmentation [1] is a fundamental task in computer vision, aiming to classify pixels in images or points in point clouds into predefined semantic categories [2]. With the advancement of sensors such as light detection and ranging (LiDAR), terrestrial laser scanning (TLS), and RGB-D cameras, point cloud data, enriched with features like color, intensity, and geometry, has become critical for tasks such as scene analysis, classification, and object detection [3]. TLS, in particular, is frequently used in environments requiring high-density 3D point cloud data, offering RGB, XYZ, and intensity channels that provide both abundant color information and accurate geometric features [4]. Due to its versatility, TLS has been widely applied to capture complex environments, such as forests, indoor spaces [5], and urban scenes. The Semantic3D dataset [6], which forms the basis of this study, is one such large-scale point cloud dataset, widely used for benchmarking segmentation algorithms.

Point cloud semantic segmentation remains a challenging task due to the sparse, unstructured, and noisy nature of point cloud data, along with occlusions caused by the 3D structure. Current approaches to point cloud segmentation generally fall into three categories: point-based, voxel-based, and image-based methods [7]. Point-based methods, pioneered by PointNet [8], directly operate on the unstructured point cloud, learning point, and global features. However, these methods often demand high computational resources and are inefficient for large-scale datasets like TLS [4]. Voxel-based methods convert the point cloud into a volumetric grid but suffer from loss of detail and reduced segmentation accuracy, particularly in environments with fine structures [9]. Image-based methods, by contrast, project 3D point clouds onto 2D images, leveraging mature 2D convolutional neural network (CNN) techniques to perform segmentation [4, 10]. This approach reduces computational complexity and achieves faster processing times, making it particularly suitable for real-time applications like autonomous driving [11]. In this study, we adopt an image-based approach using spherical projection to create panoramic images from TLS point clouds, which include RGB, intensity, and geometric information.

Although image-based methods have shown promise, especially in terms of speed and computational efficiency, they often rely heavily on RGB information. RGB data is commonly treated as a critical channel for maintaining segmentation accuracy, a perspective rooted in 2D image segmentation, where combining RGB with depth information enhances boundary delineation. However, recent research by Cai et al. [10] suggests that the inclusion of RGB channels can actually degrade segmentation accuracy when combined with other features such as intensity and geometric information. Their findings challenge the conventional assumption that RGB data is always beneficial for segmentation. They propose that inaccurate RGB information, often introduced during the colorization of point clouds, may be responsible for the reduction in accuracy. This phenomenon, although observed, has not yet been rigorously quantified, and its impact remains speculative.

Moreover, color inaccuracies in point clouds can arise due to various factors such as overexposure, sensor limitations, and motion artifacts. These errors manifest as either distinctly wrong color information or similar but incorrect color shades, both of which may adversely affect the segmentation process. However, current literature has not explored the specific effects of these inaccuracies on segmentation performance.

In this study, we address this research gap by investigating the influence of inaccurate RGB information on point cloud semantic segmentation. We propose a novel statistical

This work was supported by the Xi'an Jiaotong-Liverpool University Research Enhancement Fund under Grant REF-21-01-003.

approach to classify and quantify the impact of wrong and similar RGB information on segmentation accuracy. By comparing the results of different feature combinations, including RGB, IZeDe, and IRGBZeDe, we demonstrate how color inaccuracies disrupt segmentation, particularly in terms of geometric feature extraction. Our findings provide new insights into the role of RGB channels in image-based segmentation and offer guidance for the design of future segmentation algorithms that aim to mitigate the negative effects of color inaccuracies.

## II. METHODOLOGY

The overall experimental workflow is illustrated in Figure 1, which outlines the steps from point cloud acquisition to segmentation analysis. The process begins with preparing the Semantic3D dataset, followed by transforming the point clouds into spherical projections to create panoramic images. These images, enriched with RGB and geometric information, are used as input for the semantic segmentation network. Following segmentation, the influence of inaccurate RGB information on segmentation performance is examined by analyzing the misclassified points with respect to color errors.

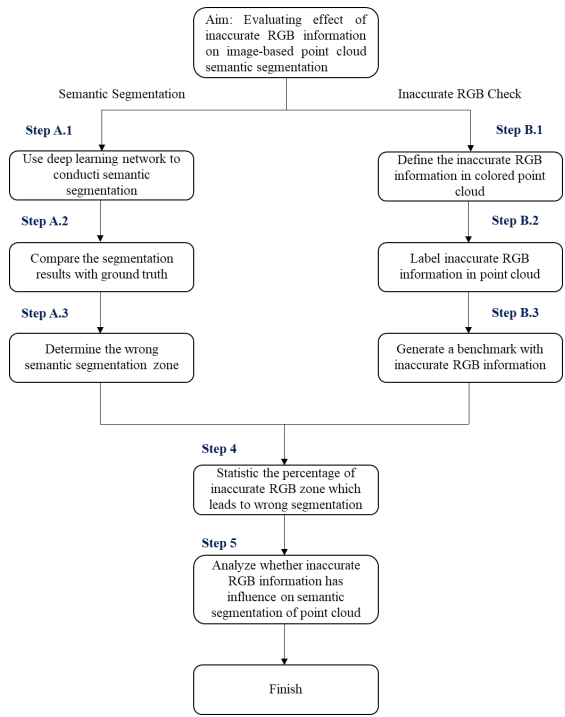

Fig. 1. Experimental procedure of point cloud segmentation and RGB error analysis.

### A. Dataset

This study utilizes the Semantic3D dataset [6], a widely used benchmark for point cloud semantic segmentation. The dataset includes 3D point cloud data captured using terrestrial laser scanning (TLS) technology, which provides high-density point clouds along with RGB, XYZ, and intensity information. Semantic3D consists of over 4 billion points, categorized into eight semantic classes: man-made terrain, natural terrain, high vegetation, low vegetation, buildings, hardscape, scanning artifacts, and cars. For this research, fifteen scenes from the dataset were selected for training, and three urban scenes were chosen as the test data: marketsquarefeldkirch-4-reduced, stgallencathedral6-reduced, and sg27_10-reduced.

Due to the fact that the Semantic3D dataset does not provide ground truth for its test set, we used the segmentation results from the state-of-the-art IZeDe channel combination [10] as a surrogate ground truth for evaluating segmentation accuracy in the test set. In the remainder of this paper, any reference to the "ground truth" of the test set refers to the segmentation results using the IZeDe channel combination.

### B. Semantic Segmentation

For semantic segmentation, the DeepLabV3+ network [12] was employed, which is a state-of-the-art deep learning model designed for high-performance image segmentation. The network includes an encoder module for extracting semantic features and a decoder module for reconstructing spatial details from the input image.

The network was initialized using pre-trained weights from the CamVid dataset [13], which contains RGB images of urban street views. However, to adapt the network to point cloud data, further training was conducted using the panoramic images generated from the Semantic3D dataset. Each image had a resolution of 2160 × 542 pixels, and the network assigned each pixel a label from one of the eight predefined categories. During this process, ground truth labels provided by Semantic3D were used for training, while the segmentation accuracy was evaluated based on the consistency between predicted labels and the ground truth.

### C. Labeling Inaccurate RGB Information

Inaccurate RGB information in point clouds can arise due to various factors such as sensor noise, overexposure, or motion artifacts during data acquisition. These inaccuracies manifest as either wrong color information, where the color is distinctly different from reality, or similar but incorrect color shades that closely resemble adjacent objects. This study categorized inaccurate RGB information into two groups, as shown in Figure 2.

Wrong RGB Information: Colors that are markedly different from the true object color, often caused by dynamic objects or exposure issues.

Similar RGB Information: Slightly different shades of similar colors, often found on object boundaries or areas with subtle color variations.

To facilitate the analysis, the inaccurate RGB points in the test data were manually labeled on a spherical point cloud with RGB information using CloudCompare software, as shown in Figure 3. The incorrect colors were marked in green (RGB values [0, 255, 0]), and similar RGB inaccuracies were labeled in yellow (RGB values [255, 255, 0]). These labeled points were then mapped back to the 2D panoramic images to assess their impact on segmentation results.

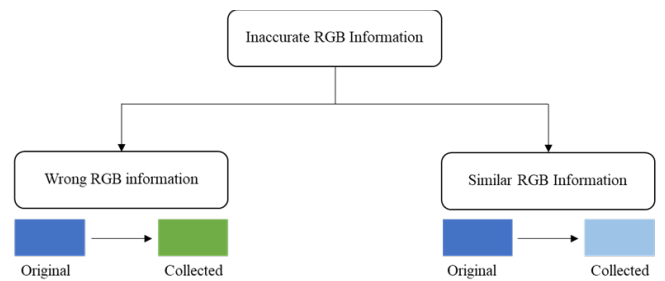

Fig. 2. The classification of inaccurate RGB information.

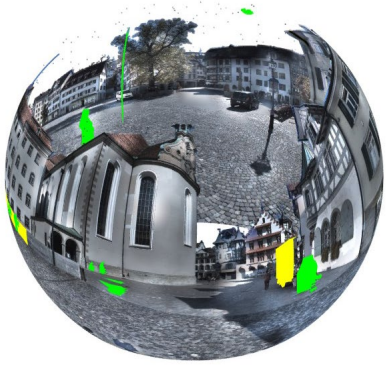

Fig. 3. Inaccurate RGB information labeled on spherical test point cloud.

### D. Data Statistics

To quantify the effect of inaccurate RGB information on segmentation performance, statistical analysis was conducted. Two main metrics were calculated:

Segmentation Accuracy: The percentage of correctly segmented pixels, calculated by comparing the predicted labels with the ground truth. The segmentation was deemed correct if the predicted RGB values matched the ground truth RGB values at the same pixel location.

Impact of Inaccurate RGB Information: The percentage of incorrect segmentation directly attributable to inaccurate RGB information. This was achieved by comparing the pixels that were incorrectly segmented with those labeled as having either wrong or similar RGB information.

For each test case, the proportion of wrong segmentation due to inaccurate RGB information was computed, allowing for an evaluation of how color inaccuracies affect segmentation performance across the different feature combinations.

## III. RESULTS

### A. Segmentation Results

Figures 4, 6, and 8 compare the semantic segmentation results between the IZeDe channels (used as ground truth) and the RGB channels for Tests 1, 2, and 3. Figures 5, 7, and 9 compare the results between ground truth and IRGBZeDe channels for the same tests. The input image size is 542 × 2160 pixels (a total of 1,170,720 pixels). In each figure, section (d) highlights the differences between the RGB-based and ground truth, with black color indicating matching pixels.

Table I summarizes the statistical results for the segmentation accuracy of both RGB-based and IRGBZeDe-based methods. The segmentation using only RGB channels achieved an average accuracy of 54.81%, and the segmentation errors are predominantly centered around occluded objects, as illustrated in Figures 4, 6, and 8. The IRGBZeDe segmentation method, which incorporates detailed geometric information, achieved an average accuracy of 95.15%. Most segmentation errors for this method occurred at object edges, as seen in Figures 5, 7, and 9.

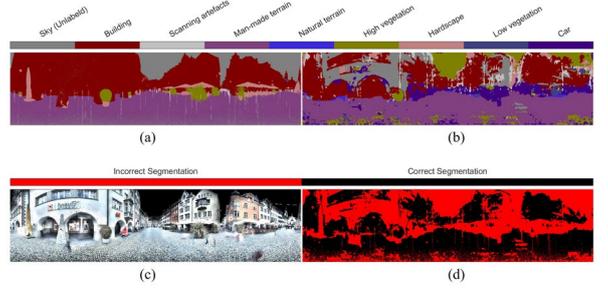

Fig. 4. Segmentation results of Test 1. (a) Ground truth. (b) Segmentation results based on RGB channel. (c) The original RGB panoramic image. (d) Segmentation difference between RGB and ground truth.

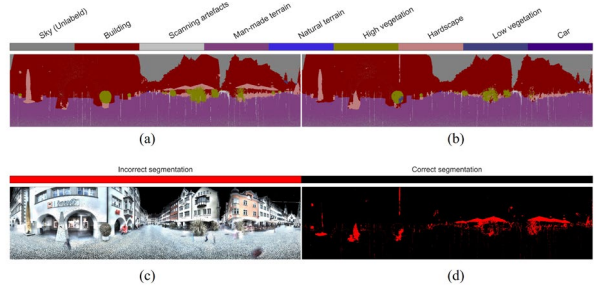

Fig. 5. Segmentation results of Test 1. (a) Ground truth. (b) Segmentation results based on IRGBZeDe channel. (c) The original RGB panoramic image. (d) Segmentation difference between IRGBZeDe and ground truth.

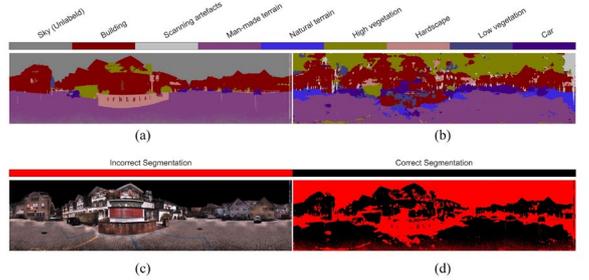

Fig. 6. Segmentation results of Test 2. (a) Ground truth. (b) Segmentation results based on RGB channel. (c) The original RGB panoramic image. (d) Segmentation difference between RGB and ground truth.

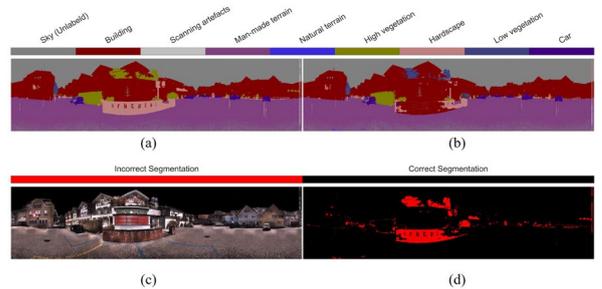

Fig. 7. Segmentation results of Test 2. (a) Ground truth. (b) Segmentation results based on IRGBZeDe channel. (c) The original RGB panoramic image. (d) Segmentation difference between IRGBZeDe and ground truth.

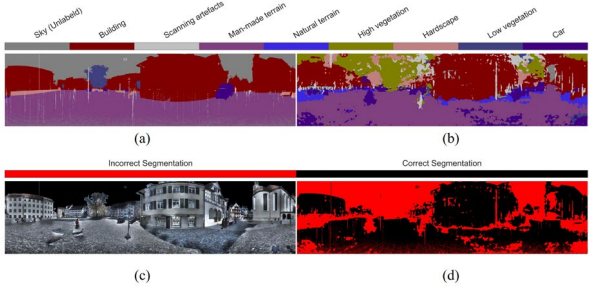

Fig. 8. Segmentation results of Test 3. (a) Ground truth. (b) Segmentation results based on RGB channel. (c) The original RGB panoramic image. (d) Segmentation difference between RGB and ground truth.

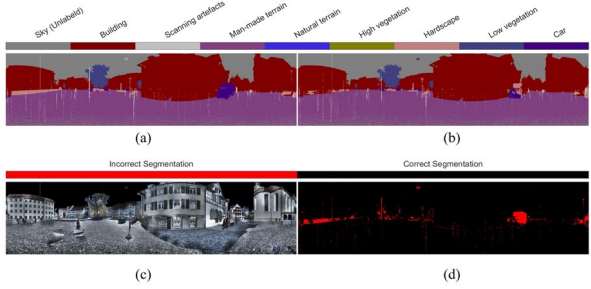

Fig. 9. Segmentation results of Test 3. (a) Ground truth. (b) Segmentation results based on IRGBZeDe channel. (c) The original RGB panoramic image. (d) Segmentation difference between IRGBZeDe and ground truth.

TABLE I. SEGMENT ACCURACY OF DEEPLABV3+ USING IRGBZEDE, RGB, AND PRETRAINED RGB.

|         | IRGBZeDe | RGB    | RGB (Pretrained) |
|---------|----------|--------|------------------|
| Test 1  | 93.30%   | 54.12% | 63.35%           |
| Test 2  | 94.35%   | 51.39% | 49.06%           |
| Test 3  | 97.79%   | 58.92% | 54.28%           |
| Average | 95.15%   | 54.81% | 54.19%           |

### B. Inaccurate RGB information

Figures 10, 11, and 12 illustrate the labeling of inaccurate RGB information across the three test datasets. In these figures, incorrect (wrong) RGB information is marked in green (RGB values [0, 255, 0]), while similar but inaccurate RGB information is marked in yellow (RGB values [255, 255, 0]). The labeled points were projected back onto the panoramic images for each test scene. This labeling was used to analyze how these color inaccuracies influence the segmentation results.

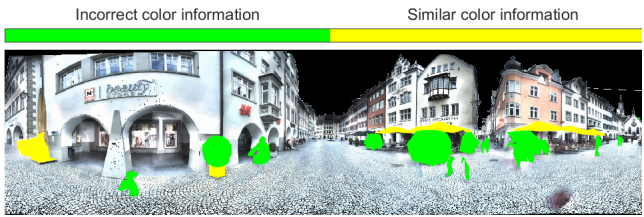

Fig. 10. Inaccurate RGB information in the Test 1 image.

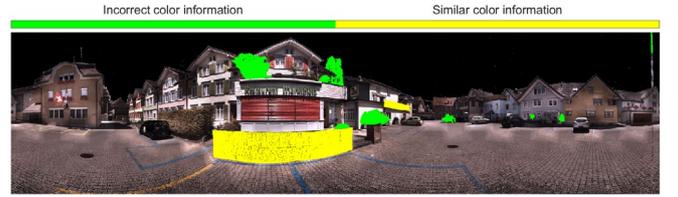

Fig. 11. Inaccurate RGB information in the Test 2 image.

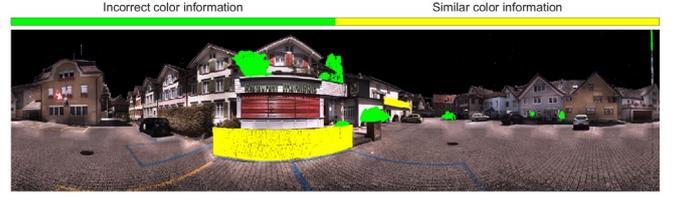

Fig. 12. Inaccurate RGB information in the Test 3 image.

### C. Statistics

After identifying the wrong and similar color information, these labeled points were matched to the corresponding segmented areas in each test. The goal was to quantify the proportion of incorrect segmentations that occurred within regions with inaccurate RGB information.

The results, as shown in Table II and Figure 13, indicate that RGB-only segmentation is more severely affected by inaccurate RGB information compared to IRGBZeDe. For RGB segmentation, wrong RGB information caused between 66.31% and 89.05% of incorrect segmentations, while similar RGB information contributed to between 46.24% and 80.54% of the errors. In contrast, the IRGBZeDe segmentation method, which incorporates geometric information, was less impacted by color inaccuracies, with the proportion of wrong segmentations caused by inaccurate RGB information being significantly lower.

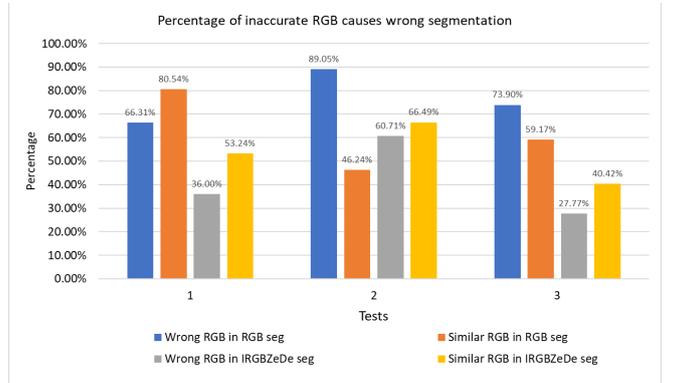

Fig. 13. Percentage of inaccurate RGB causing wrong segmentation.

TABLE II. PERCENTAGE OF INACCURATE RGB CAUSES WRONG SEGMENTATION

|        | Channel          | Wrong RGB | Similar RGB |
|--------|------------------|-----------|-------------|
| Test 1 | RGB              | 66.31%    | 80.54%      |
|        | RGB (Pretrained) | 65.26%    | 73.95%      |
|        | IRGBZeDe         | 36.00%    | 53.24%      |
| Test 2 | RGB              | 89.05%    | 46.24%      |
|        | RGB (Pretrained) | 86.19%    | 81.48%      |
|        | IRGBZeDe         | 60.71%    | 66.49%      |
| Test 3 | RGB              | 73.90%    | 59.17%      |
|        | RGB (Pretrained) | 57.13%    | 71.76%      |
|        | IRGBZeDe         | 27.77%    | 40.42%      |

The results demonstrate that inaccurate RGB information has a substantial impact on segmentation accuracy, particularly when using RGB-only segmentation. However, integrating geometric information with RGB channels in IRGBZeDe can mitigate the negative effects of color inaccuracies, especially in regions where similar color information causes segmentation errors.

## IV. Discussion and Future Work

The initial assumption was that incorrect RGB information would have a greater negative impact on segmentation than similar RGB information. This assumption holds true for RGB-only segmentation, where incorrect RGB data leads to a higher proportion of errors. However, for the IRGBZeDe method, similar RGB information caused more segmentation errors than incorrect RGB data.

This discrepancy is explained by how both types of RGB inaccuracies affect the segmentation process. Incorrect RGB data often results in misclassification in RGB-based methods by suggesting an object that doesn't fit its surroundings. In contrast, when geometric information is integrated, as in the IRGBZeDe method, adjacent points are better grouped, reducing errors from incorrect RGB data. However, similar RGB data still causes segmentation errors, particularly at object edges where both color and geometric features are important.

In summary, both incorrect and similar RGB information negatively affect segmentation accuracy when RGB channels are used. This study highlights that similar RGB data, especially when combined with geometric features, plays a significant role in misclassification.

This study shows that inaccurate RGB data, especially similar color information, significantly impacts segmentation quality, particularly around object edges. While manual labeling of incorrect RGB points was effective, future research should focus on algorithmic labeling for scalability. Developing precise algorithms for identifying color errors remains challenging due to the lack of clear definitions of "inaccurate" color.

Further research should also explore improving segmentation accuracy with RGB data. Despite the robustness of the IRGBZeDe method, RGB features can provide more detailed object segmentation if the number of labels increases. Future strategies could include:

- Multiple Scans for Data Fusion: Combining multiple scans of the same scene could reduce the impact of moving objects and improve color accuracy.
- Color Enhancement Techniques: Enhancing color contrast could mitigate the influence of similar RGB information, especially at object boundaries.
- Deep Learning for Accurate Color Identification: Training deep learning algorithms to identify and correct inaccurate color information could further improve segmentation results.

Refining the use of RGB data in segmentation models is essential for reducing errors caused by color inaccuracies and improving overall performance.

## V. Conclusion

This study explores the impact of inaccurate RGB information on point cloud semantic segmentation using three methods—RGB, IZeDe, and IRGBZeDe. It reveals that both incorrect and similar RGB data significantly degrade segmentation accuracy, particularly in RGB-only methods. While integrating geometric information in IRGBZeDe reduces the impact of incorrect RGB data, similar RGB information still causes errors, especially at object edges. The findings highlight the critical role of RGB inaccuracies in segmentation performance, emphasizing the need for future research to mitigate these effects and improve segmentation models.